\documentclass[10pt,twocolumn,letterpaper]{article}

\usepackage{cvpr}              

\usepackage{graphicx}
\usepackage{amsfonts, amsmath, amsthm, bm, amssymb}

\usepackage[figuresright]{rotating}

\usepackage{multirow}
\usepackage{makecell}
\usepackage{tablefootnote}
\usepackage{footmisc}

\usepackage[linesnumbered,ruled,vlined]{algorithm2e}
\usepackage{algpseudocode}

\SetCommentSty{mycommfont}

\newcommand{\pzero}{\phantom{0}}

\definecolor{cvprblue}{rgb}{0.21,0.49,0.74}
\usepackage[pagebackref,breaklinks,colorlinks,allcolors=cvprblue]{hyperref}

\def\paperID{13470}
\def\confName{CVPR}
\def\confYear{2025}

\title{Minimal Interaction Separated Tuning: A New Paradigm for Visual Adaptation}

\author{Ningyuan Tang, Minghao Fu, Jianxin Wu*\\
National Key Laboratory for Novel Software Technology\\
School of Artificial Intelligence, Nanjing University, China\\
{\tt\small tangny@lamda.nju.edu.cn, fumh@lamda.nju.edu.cn, wujx2001@gmail.com}}

\begin{document}
\maketitle

\renewcommand{\thefootnote}{\fnsymbol{footnote}} 
\footnotetext[1]{J. Wu is the corresponding author. This paper was partly supported by the National Natural Science Foundation of China under grant 62276123.}

\begin{abstract}
  The rapid scaling of large vision pretrained models makes fine-tuning tasks more and more difficult on devices with low computational resources. We explore a new visual adaptation paradigm called separated tuning, which treats large pretrained models as standalone feature extractors that run on powerful cloud servers. The fine-tuning carries out on devices which possess only low computational resources (slow CPU, no GPU, small memory, etc.) Existing methods that are potentially suitable for our separated tuning paradigm are discussed. But, three major drawbacks hinder their application in separated tuning: low adaptation capability, large adapter network, and in particular, high information transfer overhead. To address these issues, we propose Minimal Interaction Separated Tuning, or MIST, which reveals that the sum of intermediate features from pretrained models not only has minimal information transfer but also has high adaptation capability. With a lightweight attention-based adaptor network, MIST achieves information transfer efficiency, parameter efficiency, computational and memory efficiency, and at the same time demonstrates competitive results on various visual adaptation benchmarks.
\end{abstract}   

\section{Introduction}

Pretraining a large model \cite{mae,moco,dino,dinov2} on general-purpose data and fine-tuning it on specific downstream tasks is emerging as the mainstream methodology for computer vision tasks. Knowledge or representation learned from the pretraining makes adaptation to a target domain much easier than training from scratch. 

However, some paradoxes begin to emerge within this process or learning paradigm. One natural issue is: if the downstream dataset is small, we cannot afford to and should not store one new fine-tuned model for each downstream visual task. Parameter-efficient fine-tuning (PEFT) methods~\cite{hu2021lora,yin2023one,jia2022visual,sung2022vl,dara} are proposed to fine-tune large pretrained models with only a small set of trainable parameters, by either inserting small learnable blocks, adding few learnable tokens, or learning low-rank decomposition of extra parameters. These methods achieve competitive results on various downstream adaptation tasks with less than 1\% trainable parameters compared to full fine-tuning.

While PEFT methods partially solved the storage cost of fine-tuned models, other problems remain. As the scale of pretrained models grows with a blazing fast speed, fine-tuning or even inference itself is becoming unaffordable, especially on devices with limited computational resources. On the other hand, the need for fine-tuning and inference on such devices (\eg, personal computers, mobile phones or even cameras) rather than on the cloud AI infrastructure is increasing rapidly.

\begin{figure}
    \centering
    \includegraphics[width=\linewidth]{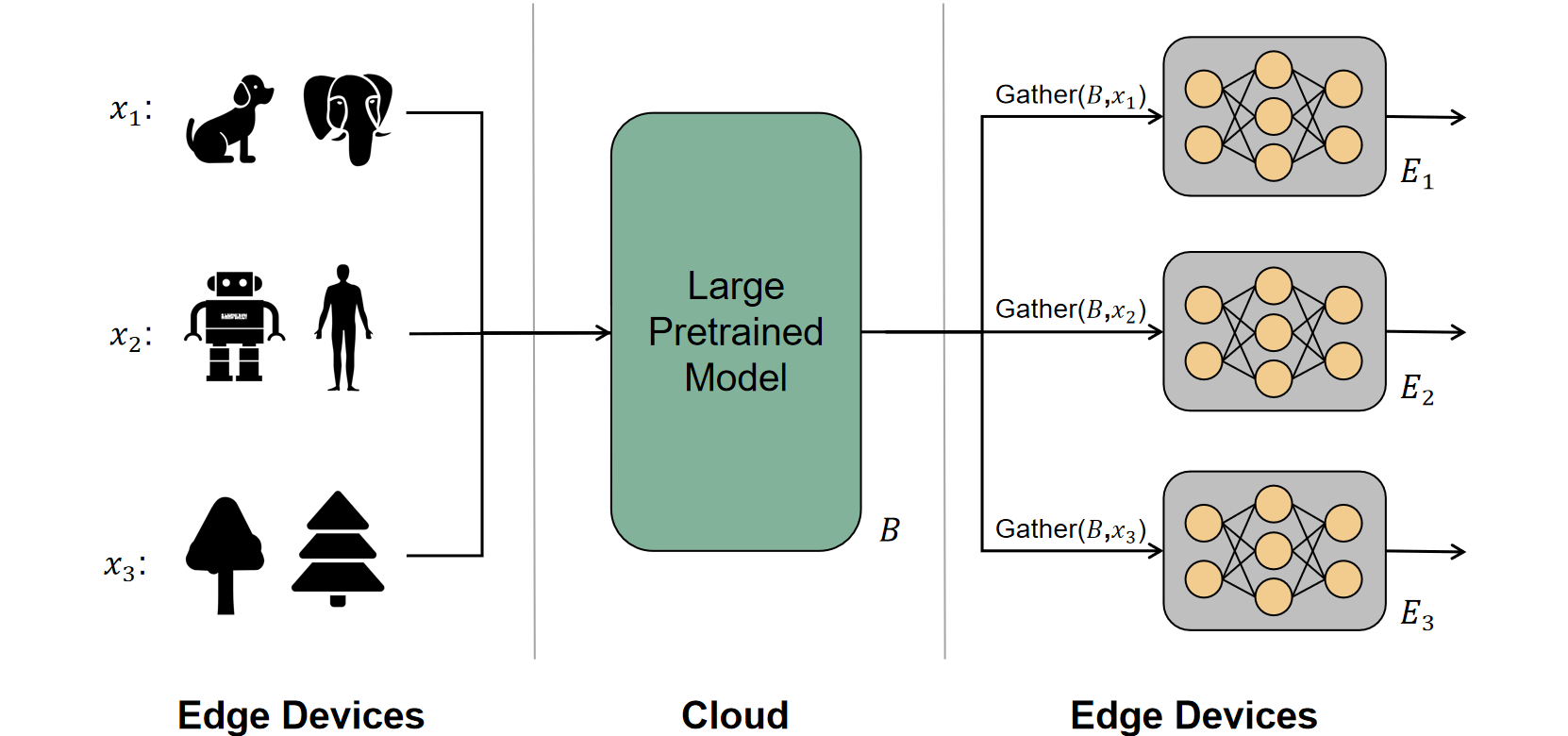}
    \caption{Illustration of our proposed Separated Tuning. Visual samples for different tasks ($x_1,x_2,x_3,\dots$) are collected and transferred to the cloud server. Within the cloud server, a large pretrained model $B$ acts as a \emph{standalone feature extractor}, producing intermediate feature sets $(B,x_i)$. A gather function compresses this set to $\mathrm{Gather}(B,x_i)$ to achieve \textbf{minimal interaction} but still keeps essential information in $(B,x_i)$ for downstream task learning. The gathered features for task $i$ is sent to a low-resource device $E_i$, where the fine-tuning is performed \emph{solely} on $E_i$ with input $\mathrm{Gather}(B,x_i)$. During inference, the pretrained model extracts features, gather them to a minimal level, transfers only a small chunk of bytes to the low-resource device, which then makes decisions with small storage, computational, and memory costs.}
    \label{fig:san}
\end{figure}

In this paper, we propose a new visual adaptation paradigm: \emph{separated tuning}, which enables low-resource devices to fine-tune downstream visual tasks with a small side network (\cf Figure~\ref{fig:san}). The side network merely receives intermediate features/activations from a pretrained model, such that the huge pretrained network is not required to be (either forward or backward) computed on low-resource devices. Upon the proposed paradigm, we design a new visual adaptation method: Minimal Interaction Separated Tuning, which is based on the side-tuning idea~\cite{sung2022lst,tang2024lowrank}. Methods in this style potentially enables separated tuning, but the \emph{interaction} (\ie, features transferred) from the pretrained network to the low-resource device is too large for a low-resource device to handle. Our contributions can be summarized as follows:
\begin{itemize}
    \item We propose a new visual adaptation paradigm: separated tuning, where the pretrained model works as a service in a cloud server, and low-resource devices fine-tune small side networks to transfer the pretrained large model towards specific downstream tasks.
    \item To reduce interaction between server and low-resource devices, we reveal that sum of intermediate features effectively capture essence of the pretrained network, which leads to both minimal interaction and high accuracy.
    \item We propose a new visual adaptation method: MIST, as an effective means for separated tuning. MIST is parameter-efficient, GPU memory-efficient, computation efficient, and enjoys low communication overhead between a cloud server and low-resource devices. Experimental results on various datasets show that MIST is comparable with state-of-the art PEFT methods in visual adaptation, but enjoys significantly smaller resource consumptions.
\end{itemize}

\section{New Pipeline: Separated Tuning}

As illustrated and described in Figure~\ref{fig:san}, the idea and structure of separated tuning are pretty simple. Computations are split between a pretrained model $B$ (residing in a cloud server) and fine-tuning modules (which usually happens in edge devices). The pretrained model $B$ is general-purpose and pretrained with abundant data and computing resources, which is too big to infer on a low-resource device. The low-resource device is mostly likely not powerful even to store the model $B$ itself, nor does it has the compute or GPU memory resources to forward compute $B$. The solution is to let $B$ extract features on cloud and transfer them to the low-resource device.

Very important but less observed or discussed, the low-resource device often lacks communication resources, too. That is, in real world it is often difficult or even impossible to transfer large chunks of data between it and the server, \eg, due to bandwidth or energy constraints. Given an input image $x$, $B$ extracts features $(B,x)$ and sends it to the low-resource device. For example, $(B,x)$ can be the set of activations of all intermediate Transformer blocks' activations, which contains tens of times more bytes to transfer than that in the input $x$ itself. In such a case, communication is too heavy a burden for the low-resource device.

Practical applications, however, are often keen to fine-tune and infer with a model for downstream tasks \emph{solely on a low-resource device} with \emph{limited storage, compute, memory, and communication capabilities}. 

\subsection{Recipes and Obstacles}

Methods that allow to split computations between a large backbone and a small side-network has appeared in the community~\cite{sung2022lst,tang2024lowrank}, where the side-network receives features from the backbone but does \emph{not} send its features back to the computation graph of the backbone. Thus, during fine-tuning, the backbone does not need to compute gradients. This split greatly reduces the compute and GPU memory costs of fine-tuning.

An even simpler recipe is linear evaluation, in which final features of the pretrained model $B$ is coupled with a simple fully connected (FC) layer. Fine-tuning is then learning the FC layer, where the transferred features are tiny, the storage, compute and memory costs are small, too. Partial-$k$ tuning unfreezes the last $k$ layers of $B$ during fine-tuning. When $k$ is small (\eg, $k=1$), it might be able to fit into low-resource devices. 

However, these methods do \emph{not} yet fit the purpose of separated tuning we propose in Figure~\ref{fig:san}. The set of 5 goals in our separated tuning (small storage/parameters, small compute, small memory footprint, small communication, high accuracy) cannot be met by existing methods. For example,
\begin{itemize}
	\item Linear evaluation and partial-$k$ tuning leads to low accuracy in recognition, because they overlook the domain gap between the general-purpose model $B$ and the downstream task. 
	\item Parameter-efficient (PEFT) methods~\cite{hu2021lora,chen2022adaptformer} often requires large compute and GPU memory during fine-tuning.
	\item Side-tuning methods~\cite{sung2022lst} produce and transfer large features $(B,x)$, and the side-network may be too big for a low-resource device.
\end{itemize}

\subsection{Minimal Interaction Separated Tuning}

\begin{figure*}
    \centering
    \includegraphics[width=0.85\linewidth]{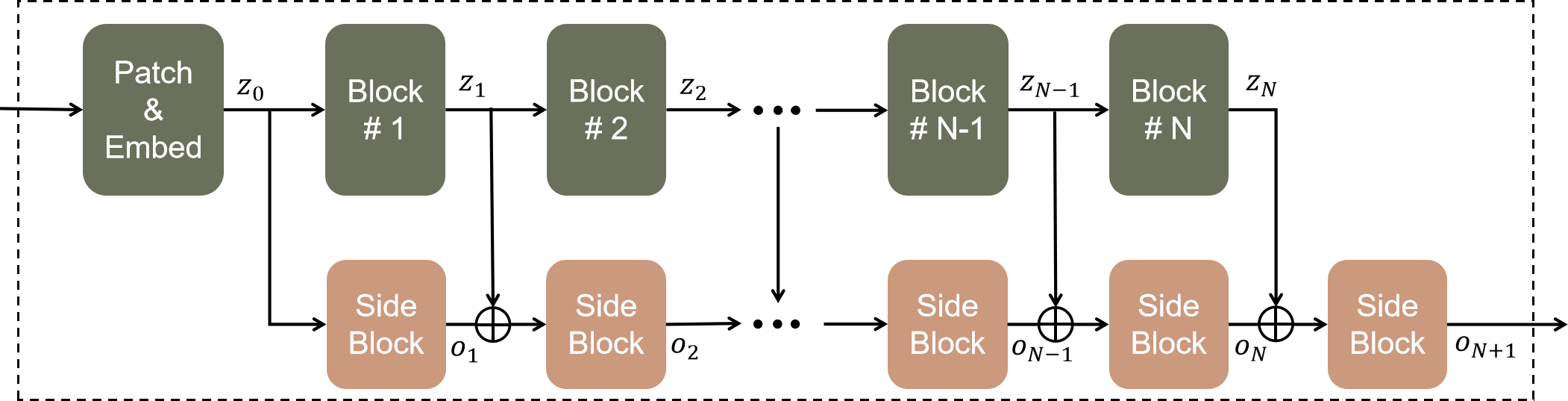}
    \caption{Architecture of a typical ladder side tuning network~\cite{sung2022lst}, where intermediate features $z_i$ from the pretrained model $B$ are added to side blocks through side paths. The interaction (features to be transferred) $(B,x)$ are the concatenation of all $z_i$ ($i=0,1,\dots,N$).
    }
    \label{fig:st_arch}
\end{figure*}

Our solution, as depicted in Figure~\ref{fig:san}, meets these requirements by proposing a gather function $\mathrm{Gather}$, which compresses the information in $(B,x)$ to $\mathrm{Gather}(B,x)$. The gathered output is succinct, hence achieving \emph{minimal interaction or communication}. So long as $\mathrm{Gather}(B,x)$ distills essential information for the downstream task, the small network on the low-resource device, $E(\mathrm{Gather}(B,x))$ will achieve high accuracy. Because in Figure~\ref{fig:san}, $E_i$ is very small and it does \emph{not} transfer any features back to $B$, the fine-tuning and inference processes resemble side-tuning methods, which are storage (parameter), compute, and memory efficient, too.

One additional benefit of our pipeline is that the pretrained model is \emph{agnostic to downstream tasks}. The pretrained model $B$ does not need to store any task-specific parameters or gradients. It simply extract features for any input images. If an input image $x$ is useful in many downstream tasks (correspondingly many low-resource devices), the pretrained model $B$ only needs to compute its features $\mathrm{Gather}(B,x)$ \emph{once}.

Then, the key and most difficult question is: how to design a gather function that \emph{captures essential information in $(B,x)$} and at the same time \emph{achieves minimal interaction}?

\section{Method}

We start by studying existing recipes. For example, side-tuning methods can be placed in our pipeline so long as we adopt the following gather function:
\begin{equation}
    \mathrm{Gather}(B,x)=\mathrm{Stack}(z_0,z_1,\cdots,z_N),
    \label{eq:big_gather}
\end{equation}
where $z_0$ is the embedding of $x$, and $z_i$ is the activations/features of the $i$-th Transformer block in $B$, as illustrated in Figure~\ref{fig:st_arch}. The `$\mathrm{Stack}$' operator stacks or concatenates all tensors $z_i$ together. For example, when the pretrained network $B$ is ViT-B, one $3\times224\times224$ input image (0.14MB in size) is required to transfer more than 7.5MB to a low-resource device!

Because side tuning methods~\cite{sung2022lst,tang2024lowrank} have achieved state-of-the-art accuracy in visual adaptation tasks, it is a natural hypothesis that $\mathrm{Stack}(z_0,z_1,\cdots,z_N)$ contains enough information for downstream tasks. And, we aim to find a new gather function $\mathrm{Gather^*}$ that can most effectively reduce its size but keep the essential information for downstream tasks.

\begin{figure*}
    \centering
    \includegraphics[width=0.8\linewidth]{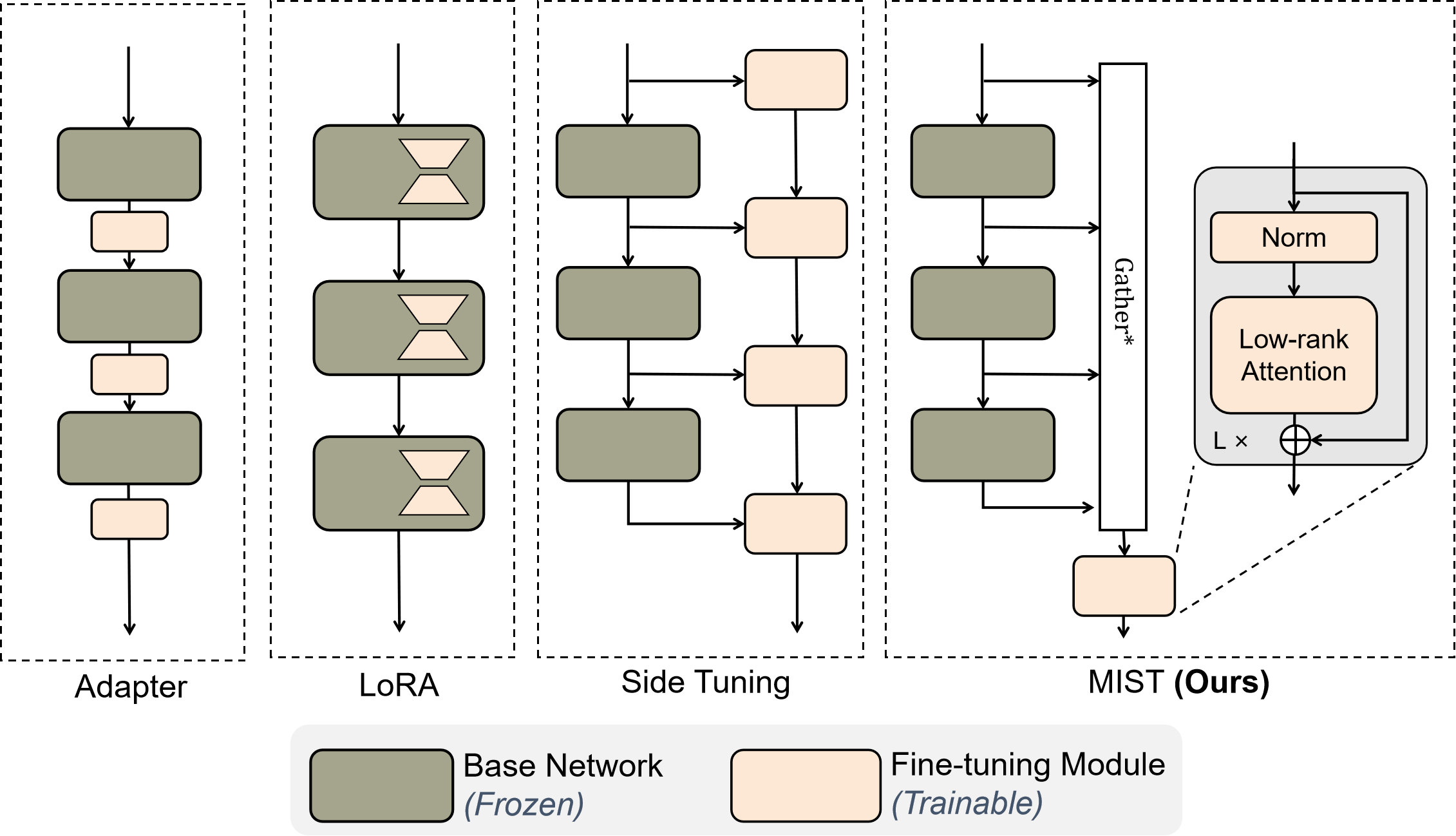}
    \caption{Illustration of MIST and other fine-tuning methods. The Adapter method fine-tunes a model by inserting trainable adapter blocks. LoRA adds trainable low-rank decomposed matrices to parameters. Side-tuning trains a decoupled side network by passing intermediate features to the side network block-wise. Our MIST compresses intermediate features by the gather function $\mathrm{Gather^*}$, and passes it to a trainable edge network to adapt to a specific task.}
    \label{fig:cmp_arch}
\end{figure*}

\subsection{Gather Function: Additive Feature Integration}
\label{ssec:sum}

Given a typical side tuning structure as shown in Figure~\ref{fig:st_arch}, learnable side blocks are placed beneath the frozen base network blocks. The $i$-th side block $S_i$ takes as input the sum of previous block output $o_{i-1}$ and corresponding feature map $z_{i-1}$ from the pretrained model $B$. Suppose the computing function in side block $S_i$ is $F_i$, then the output from $S_i$ is:
\begin{equation}
    o_i = F_i(o_{i-1}+z_{i-1}) + o_{i-1} + z_{i-1} \,.
\end{equation}
By induction, we have
\begin{equation}
     o_i = F_1(z_0)+\Sigma_{l=2}^iF_l(o_{l-1}+z_{l-1})+\Sigma_{l=0}^{i-1} z_l\,.
\end{equation}
Then the input for the $(i+1)$-th side block is $o_i+z_i$, or equivalently,
\begin{equation}
    in_{i+1}=F_1(z_0)+\Sigma_{l=2}^iF_l(o_{l-1}+z_{l-1})+\Sigma_{l=0}^{i}z_{l} \,.
    \label{eq:in}
\end{equation}
The  first two  terms in the right-hand side of Eq. \ref{eq:in} are outputs from learnable side network blocks, while the third term is the sum of intermediate features (or activations) from the pretrained model $B$. It is worth noting that the the first two terms both depend on the side-network, while the third term does \emph{not}---it is determined by $B$ alone. We name this term (or more generally a term that only depends on $B$) as an \emph{external feature term}.

To fit into our separated tuning pipeline, $B$ can \emph{only send external feature terms} to low-resource devices, a requirement only the third term $\Sigma_{l=0}^{i}z_{l}$ satisfies. This term is developed through repeatedly applying the skip connections. 

Furthermore, because side network computations ($F_i$) are low rank transformation to achieve parameter efficiency~\cite{tang2024lowrank}, a natural \emph{conjecture} is that this third term (which is determined by the complex model $B$) contains more complicated features compared to the first two terms (which are determined by the simple computations $F_i$). In other words, it is reasonable to guess that the third term, $\Sigma_{l=0}^{i}z_{l}$, contains essential information in $in_{i+1}$. Please note that this conjecture is made with respect to \emph{any} $i$ ($1 \le i \le N$), not only the final block (\ie, when $i=N$).

Our analysis in the above clearly suggests a simple gather function:
\begin{equation}
    \mathrm{Gather^*}(B,x)=\mathrm{Sum}(z_0,z_1,\cdots,z_N) \,.
\end{equation}
This gather function leads to only $1/(N+1)$ communication overhead compared to the one in Eq.~\ref{eq:big_gather}. Smaller input to low-resource devices also means that the storage, compute, and memory costs will be reduced accordingly. As our experimental results will show, $\mathrm{Gather^*}(B,x)$ does contain essential information for downstream tasks, which leads to high accuracy for them.

$\mathrm{Gather^*}(B,x)$ is very simple in its formation, and the idea to sum up intermediate features may seem naive from a first peak. We want to suggest that a simple gather function is the most effective in practice: this gather function incurs almost zero extra computation or memory cost. Furthermore, this simple and natural form is derived from careful analyses of existing methods, which is endorsed by both existing experimental results and supportive conjectures.

\subsection{Gather Function: Migrating to Downstream Tasks}

One difficulty in handling different downstream tasks are the domain gaps. In some simple tasks with only low-level image patterns, high-level features extracted by a pretrained model may be harmful for adaptation. In such cases, gathering all the intermediate features altogether is suboptimal.

So we migrate the extracted features to specific downstream tasks by only gathering features from the first $k$ blocks, which leads to the final form of our minimal interaction gather function:
\begin{equation}
    \mathrm{Gather^*}(B,x,k)=\mathrm{Sum}(z_0,z_1,\cdots,z_k) \,,
\end{equation}
in which $0 \le k \le N$. As we emphasized earlier, our hypothesis that the third term in Eq.~\ref{eq:in} contains essential information inside $in_{i+1}$ is made for all $i$ ($1 \le i \le N$). Hence, this final form is also supported by our hypothesis.

We also want to point out that previous side-tuning methods such as~\cite{sung2022lst,tang2024lowrank} correspond to $k=N$ (\ie, use the feature set $\{z_0,z_1,\dots,z_N\}$ all the way up to the final layer's output $z_N$). However, as our analysis and experiments show, it is both reasonable and beneficial to use $i<N$ for tasks with large domain gaps.

This final form introduces a hyperparameter, $k$, which needs to be set before separated tuning. We select the best $k$ from a candidate set by using the validation set. The candidate set is formed by $N$, $3N/4$ and $N/2$. For example, for ViT-B with 12 layers, we pick $k$ from the set $\{6,9,12\}$.

\begin{table*}[t]
\centering
\setlength{\tabcolsep}{1.5pt}
\small

\begin{tabular}{l|c|c|c|cc|cc|cccc}
\toprule
\textbf{ViT-B/16}
&\textbf{Support}
&\textbf{Transfer}
&\textbf{Trainable}
&\multicolumn{2}{c}{\textbf{MACs}}
&\multicolumn{2}{c}{\textbf{GPU Mem}}
&\multicolumn{4}{c}{\textbf{VTAB}}
\\
\textbf{}
&\textbf{Edge}
&\textbf{Overhead}
&\textbf{Params}
&\textbf{Cloud}&\textbf{Edge}
&\textbf{Cloud}&\textbf{Edge}
&\multicolumn{4}{c}{\textbf{}}
\\
\textbf{(85.8M)}
&\textbf{Tuning}
&\textbf{(MB/img)}
&\textbf{(M)} &
$\bm{(\times {10}^9)}$ & $\bm{(\times {10}^9)}$
&\textbf{(GB)}&\textbf{(GB)}
&\bf{Nat.} &\bf{Spe.} &\bf{Str.} &\bf{Avg.}\\
\midrule
\# of tasks &&&&&&&
&7 & 4 & 8 & 19 \\
\midrule
Full fine-tuning & & - & 85.8 & \pzero\pzero\pzero0 & 17.57 & \pzero\pzero0 & 6.09 & 75.9 & 83.4 & 47.6 & 68.9 \\
BitFit & & - & 0.10 & \pzero\pzero\pzero0 & 17.57 & \pzero\pzero0 & 3.80 & 73.3 & 78.2 & 44.1 & 65.2 \\
VPT & & - & 0.56 & \pzero\pzero\pzero0 & 23.18 & \pzero\pzero0 & 5.63 & 78.5 & 82.4 & 54.9 & 72.0 \\
LoRA & & - & 0.29 & \pzero\pzero\pzero0 & 21.87 & \pzero\pzero0 & 3.40 & 79.5 & 84.6 & 59.8 & 74.5 \\
AdaptFormer & & - & 0.16 & \pzero\pzero\pzero0 & 17.60 & \pzero\pzero0 & 4.11 & 80.6 & 84.9 & 59.0 & 74.7 \\
FacT & & - & \textbf{0.07} & \pzero\pzero\pzero0 & 18.67 & \pzero\pzero0 & 4.81 & 80.6 & 85.3 & 60.7 & 75.6 \\
SPT-LoRA & & - & 0.54 & \pzero\pzero\pzero0 & 20.44 & \pzero\pzero0 & 5.13 & \textbf{81.9} & 85.9 & 61.3 & 76.4 \\
\midrule
Linear probing & \checkmark & 0.003 & \pzero\pzero0 & 17.57 & \pzero\pzero\pzero0 & 0.57 & 0.07 & 69.1 & 77.1 & 26.8 & 57.6 \\
Partial-1 & \checkmark & \textbf{0.577} & 6.75 & 17.57 & \pzero1.42 & 0.57 & 0.52 & 69.4 & 78.5 & 34.2 & 60.7 \\
LST & \checkmark & 7.503 & 2.38 & 17.57 & \pzero0.51 & 0.57 & 2.08 & 77.6 & 85.6 & 59.7 & 74.3 \\
LAST & \checkmark & 4.040 & 0.66 & 17.57 & \pzero0.15 & 0.57 & 0.77 & 80.9 & 86.2 & 62.3 & 76.5 \\
MIST (ours) & \checkmark & \textbf{0.577} & 0.38 & 17.57 & \textbf{\pzero0.09} & 0.57 & \textbf{0.45} & 81.4 & \textbf{86.4} & \textbf{62.4} & \textbf{76.7} \\

\bottomrule

\end{tabular}

\caption{Results on VTAB-1K. ``Params'': number of trainable parameters. ``MACs Cloud'', ``MACs Edge'': multiply–accumulate operations for cloud and low-resource devices during fine-tuning. ``GPU Mem Cloud'', ``GPU Mem Edge'': GPU memory footprint within cloud and low-resource devices, respectively during fine-tuning with batch size 32. ``Nat.'', ``Spe.'', ``Str.'': average accuracy for ``Natural'', ``Specialized'', ``Structured'' task groups. ``Avg.'': group-wise average accuracy. Best results are in boldface.}
\label{tab:cmp_vtab}

\end{table*}

\subsection{The Proposed MIST Method}

With $\mathrm{Gather^*}$, our Minimal Interaction Separated Tuning (MIST) method becomes clear and easy to describe. It contains a pretrained base vision transformer network $B$ and a low-rank attention edge network $\mathrm{LAE}$ (which will be introduced latter). 

First we gather the intermediate features with $\mathrm{Gather^*}$. For base network $B$ with $N$ layers, it receives an input image $x$ and a $k$ for gather function. The forward path of $x$ produces a set of feature maps, Then we gather the feature maps with $\mathrm{Gather^*}$ to form a mixed feature map $Z_\mathrm{mix}$:
\begin{equation}
    Z_\mathrm{mix}=\mathrm{Gather^*}(B,x,k)\,.
\end{equation}
Then the low-rank attention edge network $\mathrm{LAE}$ takes $Z_\mathrm{mix}$ as input and output the final tuning result:
\begin{equation}
    Z_\mathrm{out}=\mathrm{LAE}(Z_\mathrm{mix})\,.
\end{equation}

From the result in~\cite{tang2024lowrank}, attention block with low-dimensional QKV projection is an effective and efficient structure for visual adaptation. Our experimental results also show that it is very effective for separated tuning. So we adopt this structure for edge network.

As shown in Figure~\ref{fig:cmp_arch}, $\mathrm{LAE}$ contains $L$ blocks. Each block contains a layer norm~\cite{ba2016layer} and a low-rank attention, where low-rank attention refers to self-attention with an extremely low QKV hidden dimension $r$ (\eg, hidden dimension equals 16 for feature dimension equals 768). Skip connection is also applied in each block. In practice, we take $L=4$ and $r=32$ for all downstream tasks, and find that such a small edge network is already enough for most visual adaptation tasks.

\section{Experiments}

\subsection{Experiments on VTAB-1K}

We conducted experiments on VTAB with a ViT-B/16 pretrained with ImageNet-21k. Visual Tasks Adaptation Benchmark (VTAB)~\cite{zhai2019large} is a collection of visual adaptation tasks widely used in assessing transferability of pretrained vision models and fine-tuning methods. It is a benchmark composed of 19 tasks, which can be categorized into three groups: Natural, Specialized, and Structured. Each task includes 800 training images and 200 validation images.

\paragraph{Baseline methods.} We compare MIST with various fine-tuning methods. Basically, we divide them into two categories: those potentially support separated tuning and those do not. For the first category, we have linear probing, Partial-1 (which unfreeze the final layer of ViT-B), LST~\cite{sung2022lst}, LAST~\cite{tang2024lowrank} and our method MIST. For the second category, we have BitFit~\cite{zaken2021bitfit}, VPT~\cite{jia2022visual}, LoRA~\cite{hu2021lora}, AdaptFormer~\cite{chen2022adaptformer}, FacT~\cite{jie2023fact}, SPT-LoRA~\cite{sensitivity}.

\paragraph{Results.} The results on VTAB-1K are shown in Table~\ref{tab:cmp_vtab}. Methods that potentially support separated tuning are marked with $\checkmark$ in corresponding cells. In the ``Transfer Overhead'' column, we list the communication cost per image between cloud and low-resource devices. MIST only requires to transfer 0.58MB of information per image, lower than other compared methods (except linear probing). This overhead is only about $1/7$ of LAST and $1/13$ of LST. Hence, MIST has met our ``minimal interaction'' goal.

With respect to trainable parameters, MIST only introduces 0.38M trainable parameters, lower than other methods that support separated tuning. Though MIST requires slightly more trainable parameters than some non-separated tuning methods, this difference is in fact negligible with respect to parameter size of the pretrained model (\eg, MIST: 0.38/85.8=0.44\% vs. FacT: 0.07/85.8=0.08\%, both $<1\%$).

\begin{figure*}
    \centering
    \includegraphics[width=0.6\linewidth]{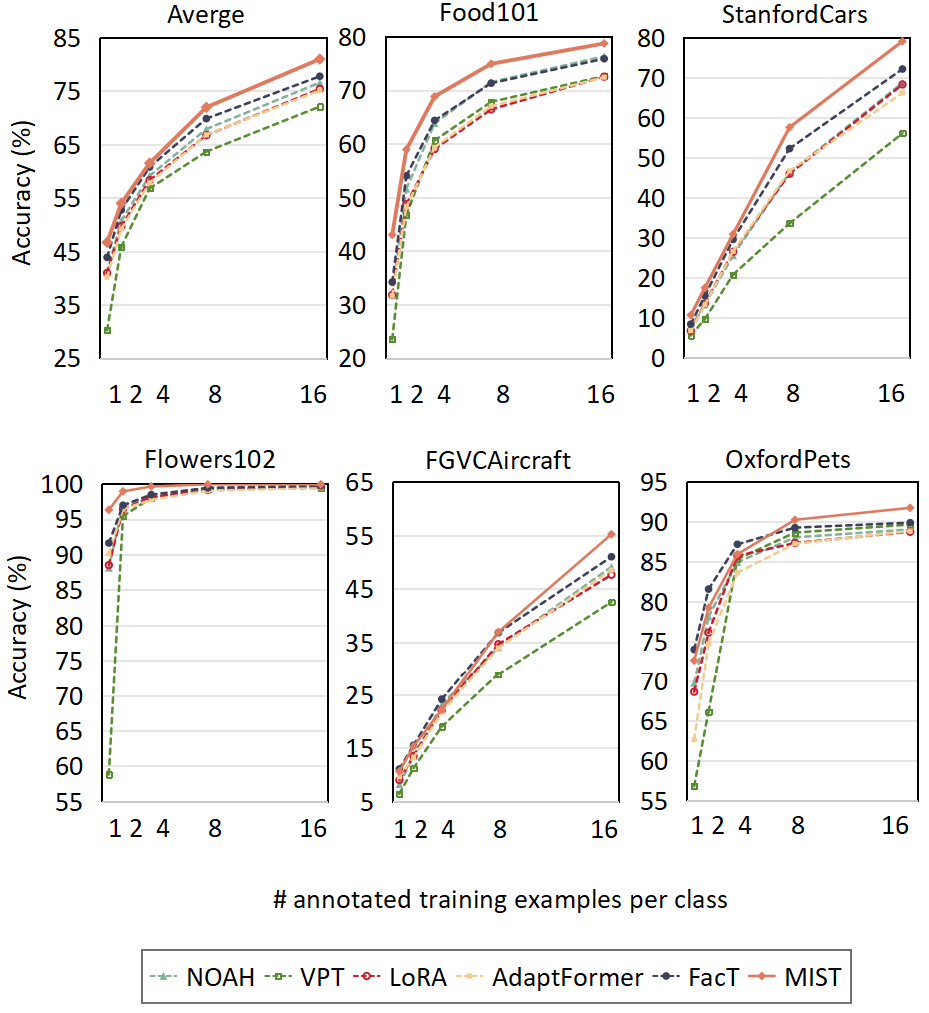}
    \caption{Top-1 accuracy on fine-grained few-shot datasets with train set containing {1, 2, 4, 8, 16}-shot per class.}
    \label{fig:fgfs}
\end{figure*}

We also list the MACs and GPU memory footprint. For non-separated tuning methods, the whole model are enforced to be fine-tuned on low-resource devices. For methods that support separated tuning, the pretrained model is deployed on a cloud server and side networks are deployed on low-resource devices, leading to very low MACs and memory footprint on low-resource devices. MIST requires the lowest MACs and GPU memory (again except linear probing) on low-resource devices, which are respectively about 1\% and 10\% of normal PEFT methods. 

MIST achieves the highest group-wise accuracy compared to other methods. It has the highest group accuracy for specialized datasets and structured datasets. This result proves that MIST has strong adaptation ability and our $\mathrm{Gather^*}$ indeed captures essential information in intermediate features.

\subsection{Experiments on Fine-Grained Few-Shot Learning}

To further assess the few-shot adaptation ability of MIST, following \cite{jie2023fact} we ran MIST on the fine-grained few-shot (FGFS) benchmark, which contains 5 fine-grained datasets: Aircraft~\cite{aircraft}, Food101~\cite{food}, Pets~\cite{dog}, Flowers102~\cite{flowers} and Cars~\cite{cars}. We fine-tuned the edge network with train sets containing \{1, 2, 4, 8, 16\}-shots per class. Experiments were repeated three times with three different random seeds, and we report the average test set accuracy.

\paragraph{Results.} As shown in Figure \ref{fig:fgfs}, the average few-shot accuracy of MIST consistently outperforms other 5 baseline methods in all cases. Average improvement of MIST compared to previous state-of-the-art, FacT, reaches 3.2\% in 16-shot, which is the greatest improvement among all settings.

\subsection{Experiments on Domain Generalization}

Following the setting in~\cite{zhang2022neural}, we tested the robustness of MIST to domain shifts~\cite{domain} when they were inevitable. We randomly sampled 16 images from each class in ImageNet-1K to form the train set. Apart from the validation set from ImageNet-1K~\cite{imagenet}, we also tested the fine-tuned model on four out-of-domain datasets: 1) ImageNet-V2~\cite{imagenet-v2} which is collected from different sources than ImageNet but following the same collection protocol, 2) ImageNet-Sketch~\cite{imagenet-sketch} which is composed of sketch images of the same 1,000 classes in ImageNet, 3) ImageNet-A~\cite{imagenet-a} which contains adversarially-filtered images, 4) ImageNet-R~\cite{imagenet-r} which contains various artistic renditions of ImageNet-1K. We repeated the experiments for three times with different seeds and report average result over three seeds.

\begin{table}
    \centering
    \small
    \begin{tabular}{l|c|cccc}
        \toprule
         & \textbf{Source} & \multicolumn{4}{c}{\textbf{Target}} \\
        \midrule
          & ImageNet & -Sketch & -V2 & -A & -R \\
        \midrule
            Adapter & 70.5 & 16.4 & 59.1 & 5.5 & 22.1 \\
            VPT & 70.5 & 18.3 & 58.0 & 4.6 & 23.2 \\
            LoRA & 70.8 & 20.0 & 59.3 & 6.9 & 23.3 \\
            NOAH & 71.5 & 24.8 & 66.1 & 11.9 & 28.5 \\
        \midrule
            MIST \textbf{(ours)} & \textbf{76.5} & \textbf{37.9} & \textbf{66.5} & \textbf{21.4} & \textbf{37.5}\\
        \bottomrule 
    \end{tabular}    
    \caption{Results on domain generalization with ViT-B/16 as the pretrained model $B$. Following~\cite{zhang2022neural}, we randomly sample 16 images from each class in ImageNet-1K, and test the fine-tuned model on test sets with different domain shifts. We report average test accuracy over 3 different random seeds. Best Results are shown in boldface.}
    \label{tab:cmp_dg}
\end{table}

\paragraph{Results.} As shown in Table~\ref{tab:cmp_dg}, MIST outperforms previous methods consistently in both the source domain and in generalization to other domains. Compared to the previous state-of-the-art method NOAH, the accuracy gain of MIST is significant in all ImageNet, ImageNet-Sketch, ImageNet-A and ImageNet-R, which are 5.0, 13.1, 9.5 and 9.0 percentage points, respectively. These result verifies the robustness of MIST against test set domain shift.

\begin{table}
    \centering
    \small
    \setlength{\tabcolsep}{3pt}
    \begin{tabular}{cc|c}
    \toprule
    $\mathrm{Gather}^*$ & LAE & Avg. \\
    \midrule
     & & 57.6 \\
    \checkmark & & 59.3 \\
    \midrule
    & \checkmark & 66.7 \\
    \checkmark & \checkmark & 76.7\\
    \bottomrule
    \end{tabular}
    \caption{Ablations on removing important components of MIST. ``Avg.'' denotes the group wise average accuracy on VTAB. $\mathrm{Gather}^*$ for adding intermediate features together or using output feature alone. LAE for using our low-rank attention edge network or linear probing).}
    \label{tab:two}
\end{table}

\begin{table}
    \centering
    \small
    \setlength{\tabcolsep}{3pt}
    
    \begin{tabular}{lc|c|c}        
    \toprule
    Block Func & \#Layers & Params (M) & Avg. \\
    \midrule
    LAE & 4 & 0.38 & 76.7 \\
    LAE & 8 & 0.76 & 76.9 \\
    \midrule
    MLP & 4 & 0.38 & 75.8 \\
    MLP & 8 & 0.76 & 76.1 \\
    \midrule
    DWConv & 4 & 0.03 & 69.8 \\
    DWConv & 8 & 0.06 & 70.2 \\
    \bottomrule
    \end{tabular}
    \makeatletter\def\@captype{table}\makeatother\caption{Ablations on different block functions with ViT-B/16 as the pretrained model. \#Layers denotes number of layers in edge network. Avg. denotes group-wise average accuracy on VTAB.}
    \label{tab:block}
 \end{table}

\subsection{Experiments on Edge Devices}

To validate the capability of our method on edge devices, we conducted the fine-tuning and inference of the ViT-B model on a Raspberry Pi 4B board with only 2GB SDRAM (\emph{with neither NPU nor GPU}). During the fine-tuning process, with a batch size of 32 and an image size of 224x224, the edge device required only 663 MB peak memory usage in the whole finetuning process. During inference, the program on the edge device required only 255 MB of memory. This level of resource consumption allows the algorithm to run smoothly on many small embedded devices.

The experiment on real world edge devices (in our case a Raspberry Pi) effectively demonstrates the capability of our method to update models in real-time on edge devices and other low-resource devices.

\subsection{Ablation studies}

\begin{figure*}
    \centering
    \includegraphics[width=\linewidth]{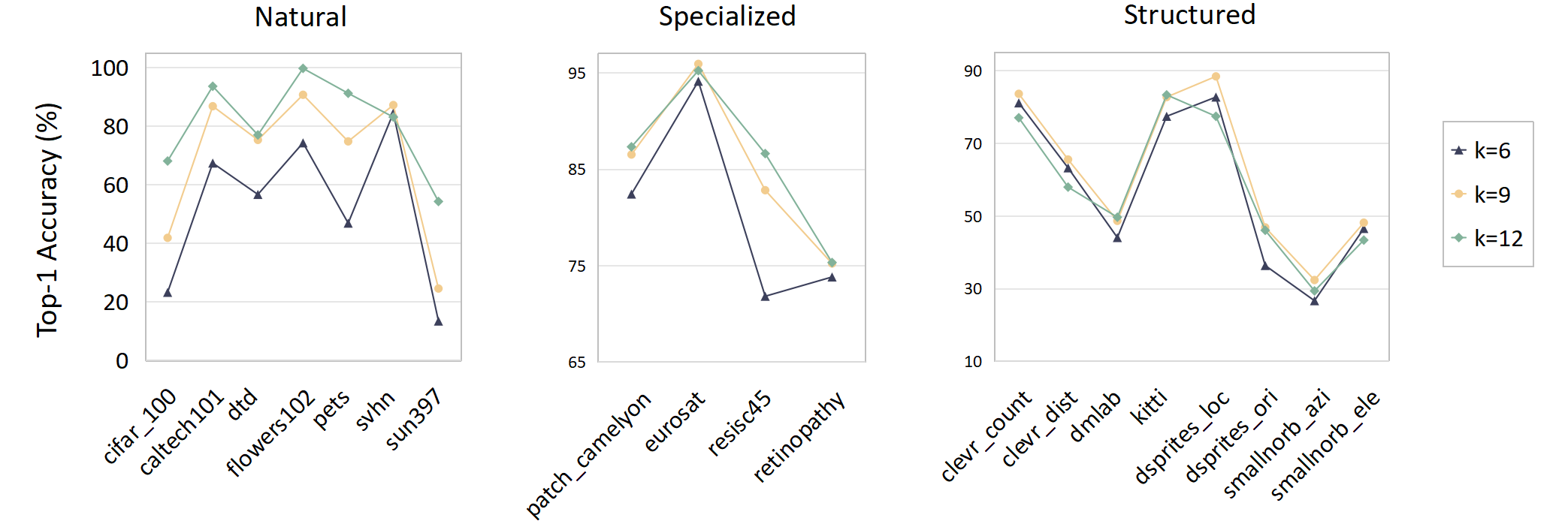}
    \caption{Ablation study on gather function with different $k$ on VTAB. Figures from left to right respectively refer to ``Natural'', ``Specialized'', and ``Structured'' task group.}
    \label{fig:gather}
\end{figure*}

We also conducted ablation studies on different components of MIST.

\paragraph{Ablations on feature gathering and LAE.} we tested the effectiveness of minimal interaction feature gathering ($\mathrm{Gather}^*$) and low-rank edge network in our method. Results are shown in Table~\ref{tab:two}. When both are not adopted, the method degrade to naive linear probing. When only minimal interaction feature gathering is used, then intermediate features are added for linear probing. The result in the second row shows a 1.7\% gain over naive linear probing, but this improvement is limited. When only LAE is used, the accuracy gain over naive linear probing is 9.1\%, which indicates that adding an edge network directly after a pretrained model is useful for visual adaptation, but still lags far behind other PEFT methods. When both $\mathrm{Gather}^*$ and LAE are adopted, the average accuracy increases to 76.7\% (on par with or better than other fine-tuning methods). This ablation shows that both components are important for MIST in visual adaptation, and only when two components are used together can we get the best results.

\paragraph{Ablations on different edge network structure.} We substituted our LAE block function in the edge network with MLP or Depthwise convolution (DWConv)~\cite{dwconv}. Results are shown in Table~\ref{tab:block}. We can see that while low-rank attention (LAE) and MLP have the same number of parameters, LAE outperforms MLP with about 1\% average accuracy lead. When the block function is changed to depthwise convolution, the number of parameters decreases, but the average accuracy also drops by a large margin. Overall, low-rank attention leads to the best performance and parameter efficiency tradeoff.

\paragraph{Ablations on domain migration.} To verify the effect of domain migration with different $k$, we tested the respective accuracy on all 19 VTAB datasets for $k=6$, $k=9$ and $k=12$. Results are shown in Figure~\ref{fig:gather}. For Natural and Specialized tasks, $k=12$ (adding all intermediate features) usually lead to the best result. But for `Structured' which have more low-level features, $k=9$ outperforms $k=12$ in 6 out of 9 datasets. These results indicate that in datasets with rich low-level features, intermediate outputs from early layers of the pretrained model are more important.

\section{Related Work}

Parameter-efficient fine-tuning (PEFT) aims at fine-tuning a large pretrained model with only a small amount of trainable parameters. Adapter~\cite{houlsby2019parameter} inserts learnable MLP layers after each attention~\cite{vaswani2017attention} layer in a Transformer block. AdaptFormer~\cite{chen2022adaptformer} inserts MLP layers to FFN in a parallel form. BitFit~\cite{zaken2021bitfit} fine-tunes the pretrained model to downstream tasks by learning all bias terms within the pretrained model.  SSF~\cite{lian2022scaling} learns scale factor $\gamma$ and shift factor $\beta$, and $X'=\gamma\odot X+\beta$, where $\odot$ is element-wise product. VPT~\cite{jia2022visual} does visual adaptation by learning additional task-specific tokens. VPT-Shallow only inserts learnable tokens before the first later, while VPT-Deep inserts learnable tokens before each later.

LoRA~\cite{hu2021lora} fine-tunes low-rank decomposition matrices of a dense layer instead of fine-tuning dense layers directly. Weights are updated with $W'=W+AB$, where $A\in\mathbb{R}^{h\times h'}$, $B\in\mathbb{R}^{h'\times h}$, $h'\ll h$. NOAH~\cite{zhang2022neural} combines the structure of Adapter, VPT and LoRA and use evolutionary algorithm to search for best combination.

Another line for visual adaptation is about memory-efficiency. LST~\cite{sung2022lst} proposes to fine-tune a side network which is disentangled from pretrained model, and save memory by freeing pretrained model from back-propagation. DTL~\cite{fu2024dtl} considers adding side information back to pretrained network to make a tradeoff between adaptation ability and memory consumption. LAST~\cite{tang2024lowrank} proposes a low-rank attention side network for effective visual adaptation. MIST stems from this line of work, but elaborates on minimizing the interaction between backbone and side networks, which previous side tuning methods (like LAST) were not capable of, but is the key to realizing the proposed `Separated Tuning'.

Another related work is Offsite-Tuning~\cite{offsite}. Offsite-Tuning extracts a small version of the base model called `emulator' to emulate the behavior of the base model, and fine-tune adapter with the emulator rather than the base model. This helps to reduce computation overhead during fine-tuning and protect data privacy.

\section{Conclusions and Discussions}

In this paper, we proposed a visual adaptation paradigm and a method which allows devices with low computational resources to finetune large vision model without access to the pretrained model itself. While a pretrained model resides in cloud and provides features that are too large to transfer to the low-resource device, our minimal interaction separated tuning (MIST) compresses the features while keeping essential information for downstream tasks. The entire fine-tuning process is performed only on the low-resource device, requesting only limited storage, compute, memory, and communication resources while achieving high accuracy. In particular, we emphasized the communication factor, which has been largely ignored in previous research.

Currently, experiments on MIST are mostly restricted to visual recognition tasks, and this might constrain the scope of usage of MIST. However, we have adapted MIST to segmentation. And the results showed that MIST is also suitable for higher level visual tasks (\cf appendix for results). More vision tasks, such as image and video generation, may also enjoy benefits brought up by MIST. We will also explore other gather functions in the new minimal interaction separated tuning framework, which may provide even better overall performance.

\clearpage
\setcounter{page}{1}
\maketitlesupplementary

\section{Detailed results on VTAB-1K}

Here we present a detailed version of Table~\ref{tab:cmp_vtab_de}, where per-task accuracy on VTAB-1K datasets are listed.

\section{Segmentation result on ADE-20k}

For semantic segmentation, we fine-tune a SegFormer-B4 model on ADE20K dataset. SegFormer has a multi-stage architecture, where the backbone is divided into 4 stages. To ensure compatibility for feature addition, we use the downsampling module in each stage to downsample the spatial dimensions of the aggregated features, but without training. This aggregated feature is used as the final feature from backbone and fine-tuned with LAE. Due to the computational limitation, we only fine-tune 42k iters for all methods, which is roughly 1/4 of the 160k iters mentioned in segformer paper. MIST achieves 2.4 mIoU gain compared to fine-tuning with frozen encoder (42.9 vs. 40.5), and is only 0.5 point behind full fine-tuning (42.9 vs. 43.4). This result shows that MIST also work

\begin{table}
    \centering
    \small
    \setlength{\tabcolsep}{3pt}
    \makeatletter\def\@captype{table}\makeatother\caption{Ablations of semantic segmentation. On ADE20K dataset with SegFormer-b4. We use \textbf{mIoU} as evaluation metric. \textbf{Params} denotes number of trainable parameters in SegFormer encoder. \textbf{Gather}$^*$ for (add intermediate features together / use output feature alone). \textbf{LAE} for (use low-rank attention edge network / not use).}
    \begin{tabular}{l|cc|c|c}
    \toprule
    \textbf{Method} & \textbf{Gather}$^*$ & \textbf{LAE} & \textbf{Params.} & \textbf{mIoU} \\
    \midrule
    Frozen & - & - & \pzero0.0M & 40.5 \\
    Full-FT & - & - & 60.8M & \textbf{43.4} \\
    \midrule
    MIST & \checkmark & & \pzero0.0M & 40.7 \\
    MIST & & \checkmark & \textbf{\pzero0.7M} & 42.3 \\
    MIST & \checkmark & \checkmark & \textbf{\pzero0.7M} & 42.9 \\
    \bottomrule
    \end{tabular}
    \label{tab:one}
\end{table}

\begin{table*}
        \footnotesize
         \centering
         \setlength{\tabcolsep}{2.5pt}
	\caption{Per-task results on the VTAB-1K benchmark with ViT-B/16 pretrained on IN21k. ``Mean Acc'' denoted groupwise average accuracy on three VTAB task groups.}
         \begin{tabular}{l|ccccccc|cccc|cccccccc|ccc}
         \toprule[1.5pt]
			& \multicolumn{7}{c|}{\textbf{Natural}} & \multicolumn{4}{c|}{\textbf{Specialized}} &  \multicolumn{8}{c|}{\textbf{Structured}} \\ 
			  & \rotatebox{90}{CIFAR-100} & \rotatebox{90}{Caltech101} & \rotatebox{90}{DTD} & \rotatebox{90}{Flowers102} & \rotatebox{90}{Pets} & \rotatebox{90}{SVHN}  & \rotatebox{90}{Sun397} & \rotatebox{90}{Patch Camelyon~} & \rotatebox{90}{EuroSAT}   & \rotatebox{90}{Resisc45}  & \rotatebox{90}{Retinopathy} & \rotatebox{90}{Clevr/count} & \rotatebox{90}{Clevr/distance}  & \rotatebox{90}{DMLab} & \rotatebox{90}{KITTI/distance~}  & \rotatebox{90}{dSprites/loc} & \rotatebox{90}{dSprites/ori}   & \rotatebox{90}{SmallNORB/azi~}  & \rotatebox{90}{SmallNORB/ele~} & \rotatebox{90}{Mean Acc} \\
			\midrule
			Linear probing & 64.4 & 85.0 & 63.2 & 97.0 & 86.3 & 36.6 & 51.0 & 78.5 & 87.5 & 68.5 & 74.0 & 34.3 & 30.6 & 33.2 & 55.4 & 12.5 & 20.0 & \pzero9.6 & 19.2  & 57.6 \\
			Full finetuning & 68.9 & 87.7 & 64.3 & 97.2 & 86.9 & 87.4 & 38.8 & 79.7 & 95.7 & 84.2 & 73.9 & 56.3 & 58.6 & 41.7 & 65.5 & 57.5 & 46.7 & 25.7 & 29.1 & 68.9 \\
                BitFit & 72.8 & 87.0 & 59.2 & 97.5 & 85.3 & 59.9 & 51.4 & 78.7 & 91.6 & 72.9 & 69.8 & 61.5 & 55.6 & 32.4 & 55.9 & 66.6 & 40.0 & 15.7 & 25.1 & 65.2 \\
                VPT & 78.8 & 90.8 & 65.8 & 98.0 & 88.3 & 78.1 & 49.6 & 81.8 & 96.1 & 83.4 & 68.4 & 68.5 & 60.0 & 46.5 & 72.8 & 73.6 & 47.9 & 32.9 & 37.8 & 72.0 \\
                LST & 59.5 & 91.5 & 69.0 & 99.2 & 89.9 & 79.5 & 54.6 & 86.9 & 95.9 & 85.3 & 74.1 & 81.8 & 61.8 & 52.2 & 81.0 & 71.7 & 49.5 & 33.7 & 45.2 & 74.3 \\
                LoRA & 67.1 & 91.4 & 69.4 & 98.8 & 90.4 & 85.3 & 54.0 & 84.9 & 95.3 & 84.4 & 73.6 & 82.9 & 69.2 & 49.8 & 78.5 & 75.7 & 47.1 & 31.0 & 44.0 & 74.5 \\
                AdaptFormer & 70.8 & 91.2 & 70.5 & 99.1 & 90.9 & 86.6 & 54.8 & 83.0 & 95.8 & 84.4 & 76.3 & 81.9 & 64.3 & 49.3 & 80.3 & 76.3 & 45.7 & 31.7 & 41.1 & 74.7 \\
                FacT & 70.6 & 90.6 & 70.8 & 99.1 & 90.7 & 88.6 & 54.1 & 84.8 & 96.2 & 84.5 & 75.7 & 82.6 & 68.2 & 49.8 & 80.7 & 80.8 & 47.4 & 33.2 & 43.0 & 75.6 \\
                SPT-LoRA & 73.5 & 93.3 & 72.5 & 99.3 & 91.5 & 87.9 & 55.5 & 85.7 & 96.2 & 75.9 & 85.9 & 84.4 & 67.6 & 52.5 & 82.0 & 81.0 & 51.1 & 30.2 & 41.3 & 76.4 \\
                LAST & 66.7 & 93.4 & 76.1 & 99.6 & 89.8 & 86.1 & 54.3 & 86.2 & 96.3 & 86.8 & 75.4 & 81.9 & 65.9 & 49.4 & 82.6 & 87.9 & 46.7 & 32.3 & 51.5 & 76.5 \\
			\midrule
                MIST & 67.8 & 93.2 & 76.8 & 99.7 & 91.1 & 87.5 & 54.0 & 87.3 & 95.8 & 87.1 & 75.3 & 84.5 & 66.8 & 51.2 & 83.7 & 84.0 & 45.7 & 33.3 & 50.3 & 76.7 \\
			\bottomrule
		\end{tabular}
	\label{tab:cmp_vtab_de}
\end{table*}

\section{Imlementation details}

We conducted all visual adaptation tasks with a ViT-B/16 pretrained on IN21k as backbone. For VTAB-1K, FGFS and domain generalization tasks, we used Adam optimizer with batch size 32, and cosine learning scheduler to adjust learning rate. Learning rate is selected according to the accuracy on validation set. Following the setting of \cite{jia2022visual}, images were directly resized to 224 × 224 for VTAB-1K. For FGFS and domain generalization tasks, following the setting in \cite{jie2023fact} and \cite{zhang2022neural}, we performed random resize crop and random horizontal flip, and no extra data augmentation tricks were used.

All the experiments were implemented with PyTorch~\cite{pytorch}.

\section{Experiment on varying side input for side tuning}

In section~\ref{ssec:sum}, we conjecture that the third term in Eq.~\ref{eq:in} ($\Sigma_{l=0}^{i}z_{l}$, or external feature term), contains essential information in $in_{i+1}$, and thus important to the fine-tune accuracy.
 So we vary the side input term, by taking $\hat{z_{i}}$ instead of $z_i$ as auxiliary input for side block $S_{i+1}$, where:

\begin{equation}
    \hat{z_{i}}=\left\{
    \begin{aligned}
        &z_i & ,i<g \\
        &z_i-z_{i-g} & ,i\geq g
    \end{aligned}
    \right.
\end{equation}

$g$ varies from 1 to N+1. then, $in_{i}=F_1(z_0)+\Sigma_{l=2}^iF_l(o_{l-1}+z_{l-1})+\Sigma_{l=0}^{i}\hat{z}_{l}$, making:

\begin{equation}
    in_{i}=\left\{
    \begin{aligned}
        &F_1(z_0)+\Sigma_{k=2}^iF_k(o_{k-1}+z_{k-1})+\Sigma_{k=0}^{i}z_{k} & ,i<g \\
        &F_1(z_0)+\Sigma_{k=2}^iF_k(o_{k-1}+z_{k-1})+\Sigma_{k=i-g+1}^{i}z_{k} & ,i\geq g
    \end{aligned}
    \right.
\end{equation}

In the experiment, we use ViT-B~\cite{dosovitskiy2020image} as backbone, and adopts the low-rank attention structure from \cite{tang2024lowrank} as block function in side network. We test the effect of different $g$ on VTAB benchmark, and results are shown in lower-right of Figure \ref{fig:st_arch} (Notice that when $g=13$, the setting falls backs to normal side tuning). From this experiment, we can find that, without previous external feature term from skip connection (the case when $g=1$), overall accuracy suffers from a big cut (73.1). So $z_i$ from the side path alone is not enough, the accumulated external feature term is the key for side network. Furthermore,  we can discover that increase of $g$ will lead to a higher accuracy, this indicates that more complicated external feature term can result in better model accuracy. The overall accuracy reaches a platform when $g\geq6$, indicating the benefit of more complicated external feature term is limited.

\begin{figure*}
    \centering
    \includegraphics[width=\linewidth]{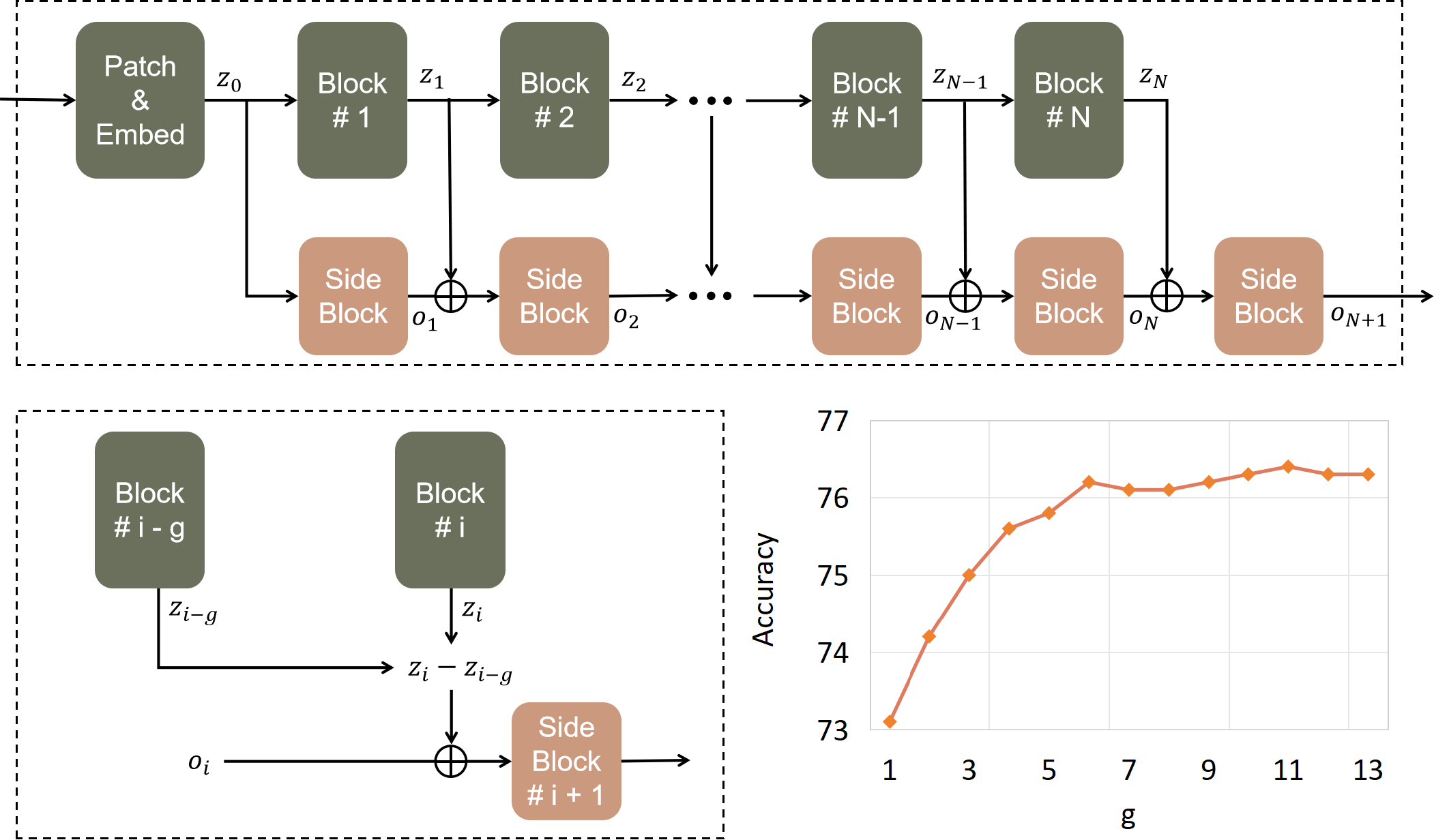}
    \caption{The upper figure shows the architecture of typical ladder side tuning, where intermediate features from pretrained model are added to side blocks through side paths. The lower-left figure shows a modified version of ladder side tuning, where the input from $i$-th side path $z_i$ is replaced with $z_{i}-z_{i-g}$. Experiment result on varying intermediate feature input to side network is shown in the lower-right figure. Where $g$ is the hyperparameter which controls the form of intermediate feature input to each side block. Accuracy stands for group-wise average accuracy on VTAB benchnark.
    }
    \label{fig:st_arch_ch}
\end{figure*}


\end{document}